\documentclass[10pt,a4paper]{article}

\usepackage{amssymb}
\usepackage{booktabs}
\usepackage{graphicx}
\usepackage{xspace}
\usepackage{amsmath}
\usepackage{amssymb}
\usepackage{amsthm}
\usepackage{balance}
\usepackage{relsize}
\usepackage{graphics}
\usepackage{subfig}
\usepackage{tikz}
\usepackage{multirow}
\usepackage{url}

\usepackage[affil-it]{authblk}
\usepackage{bm}

\mathchardef\mhyphen="2D
\makeatletter
\DeclareRobustCommand\onedot{\futurelet\@let@token\@onedot}
\def\@onedot{\ifx\@let@token.\else.\null\fi\xspace}

\def\eg{\emph{e.g}\onedot} 
\def\ie{\emph{i.e}\onedot}

\def\etal{\emph{et al}\onedot}
\makeatother

\newcommand{\myparagraph}[1]{\paragraph{\textbf{#1}}}

\newcommand{\ul}[1]{\underline{#1}}

\title{What is the right way to represent document images?}
\author{Gabriela Csurka, Diane Larlus, Albert Gordo and Jon Almaz\'{a}n}
\affil{Xerox Research Centre Europe, 6 chemin de Maupertuis \\
38240,  Meylan France \\
\texttt{Firstname.Lastname@xrce.xerox.com}}

\begin{document}
\maketitle

\begin{abstract}
{\em 
In this article we study the problem of document image representation based on visual features.  
We propose a comprehensive experimental study that compares three types of
visual document image representations: (1) traditional so-called shallow features, 
such as the RunLength and the Fisher-Vector descriptors, (2) deep features based on
Convolutional Neural Networks, and (3) features extracted from hybrid architectures that take inspiration from the two previous ones.

We evaluate these features in several tasks (\ie classification, clustering, and retrieval) 
and in different setups (\eg domain transfer) using several public
and in-house datasets.  Our results show that deep features generally outperform other 
types of features when there is no domain shift and the new task is closely related  to the one used to train the model. However, when a large
domain or task shift is present, the Fisher-Vector shallow features generalize better and often obtain the best results.}
\end{abstract}

\section{Introduction}
\label{sec:intro}

In this work we focus on the problem of document image representation and understanding.
Given images of documents, we are interested in learning how to represent the documents 
to perform tasks such as classification, retrieval, clustering, etc.
Document understanding is a key aspect in, for instance, digital mail-room scenarios, 
where the content of the documents are used to route incoming documents to the right workflow, 
extract relevant data, annotate the documents with additional information such as priority or relevance, etc.

Traditionally, there has been three main cues that are taken into account when looking to represent and understand a document image: visual cues, structural cues, and textual cues \cite{ChenBlostein07}.
The visual cues describe the overall appearance of the document, and capture the information that would allow one to differentiate documents ``at a glance''. 
The structural cues explicitly capture the relation between the different elements of the documents, for example by performing a layout analysis and encoding the different regions in a graph. Although visual descriptors can capture similar information implicitly, the representations based on structural cues focus on capturing them in an explicit manner.
Finally, textual cues capture the textual information of the document, which can contain important semantic information.

In many cases, these cues contain complementary information, and their combined use would be desired.
Unfortunately, obtaining structural and textual features is usually computationally expensive, and these costs become prohibitive in large scale domains. For
example, structural features usually require a layout analysis of the document, which is slow and error prone. Similarly, textual cues usually require to
perform OCR on the entire document, which is once again slow and error prone. Moreover, these two kinds of features are very domain-specific, and, in general,
do not transfer well between different domains and tasks.

On the other hand, visual features are usually fast to obtain while being quite generic. This has motivated their use in many document understanding works
\cite{CHH1997,HDRT1998,SD2001, BW2003, S2006,ChanChang01,KSB2007, GordoPerronninICPR10,GordoDAS12,Rusnol12,GPV13, GordoICDAR13, Rusnol14}.  Although not as
expressive as pure structural features, visual features can typically encode some coarse structure of the image, while textual information can be added as an
additional step depending on the specific domain \cite{Gordo13}.  Recently, deep learning techniques have been used to build visual representations of
documents, showing promising results and outperforming handcrafted, shallow visual features in classification and retrieval tasks \cite{lekang14,harleyetal15}.

Motivated by their success and advantages, in this work we focus on visual features for document representation, 
and propose a comprehensive experimental study where we compare 
handcrafted, shallow features with more recent, learned features based on deep learning.
In particular, although deep features have shown outstanding performance in many computer vision tasks, 
only a few works have focused on learning features for document images using convolutional networks  
(e.g. \cite{lekang14,harleyetal15}), and their comparison with other shallow methods has been limited. 
This recent shift towards deep learning in document image understanding raises two questions:
First, given a task and a dataset, do these deep methods outperform shallow features in all the cases? 
Second, how well do they transfer to different domains
(\ie datasets) and to different tasks, if one wants to reduce their training cost by reusing a pre-trained representation or model? 
 These crucial questions have not been addressed in detail yet.

Additionally, some hybrid architectures have recently been proposed for
natural image classification \cite{perronninlarlus15}. Built on top of shallow
features, they also include several layers that allow them to be trained
end-to-end similarly to deep models. The underlined motivation is to combine the
advantages of shallow features (faster training and good generalization) with
the expressiveness of deep architectures. In this paper we propose to evaluate them in the context
of document image understanding in comparison with shallow features and deep convolutional networks. 

Our contribution is therefore threefold:
\begin{itemize}
\item First, we benchmark several standard features against different flavors of recently proposed deep features on the document
classification task. 
\item Second, we explore hybrid architectures as an appealing comprise between reusable but weaker shallow features, and specialized but
high-performing deep features.  
\item Third, we evaluate the transferability of all these features across domains and across tasks.
\end{itemize}

Accordingly, this article is organized as follows. In Section \ref{sec:rw}, we review related work. Section \ref{sec:representation} describes the different
feature representations that we consider for this work. Section \ref{sec:training} details the training procedure with and without domain shift. 
Section \ref{sec:eval} describes the datasets and implementation details used in our experiments.
The experimental results in Section \ref{sec:exp} are divided into two. The first set of experiments (Section \ref{sec:part1}) compares all the features with a standard
protocol to tackle document image classification. The second set of experiments (Section \ref{sec:part2}) studies transferability of the features on different
datasets and different tasks. Finally, Section \ref{sec:ccl} concludes our benchmark study.

\section{Related Work}
\label{sec:rw}
Traditional visual features for document images usually rely on simple statistics computed directly from the image pixels.
For example, Heroux \etal  \cite{HDRT1998} propose a multi-scale density  decomposition of the  page 
 to produce fixed-length descriptors  constructed efficiently from integral images. A similar idea is presented by 
 Reddy and Govindaraju \cite{Reddy08}, where representations based on low-level pixel density information are classified using adaptive boosting.
Cullen \etal \cite{CHH1997} propose to use a combination of features including densities at interest points, 
histograms of the size and the density of the connected components and vertical projection histograms.
Bagdanov and Worring \cite{BW2003} propose a representation based on density changes obtained with different 
morphological operations, while Sarkar \cite{S2006} describes document images as a list of salient Viola-Jones based features. 
Joutel \etal \cite{joutel07} propose the use of curvelets to capture information about handwritten strokes in the image. 
However, this descriptor is tailored to the specific task of retrieving images with similar handwriting styles, 
and their use beyond that particular task is limited.

Some more elaborate representations, such as the RunLength histograms~\cite{ChanChang01,KSB2007,GPV13}, 
have shown to be more generic and hence better suited for document image representation.
Many of these representations can be combined with spatial pyramids \cite{LSP06} 
to explicitly add a coarse structure, leading to higher accuracies at the cost of higher-dimensional representations.  
However, in general, all these traditional features contain relatively limited amount of 
information and while they might perform well on a specific dataset and task for which they were designed, 
they are not generic enough to be able to handle various document class types, datasets and tasks.

On a different direction, some more recent works~\cite{GordoICDAR13,Gordo13, kumar14,docimgArXiv16} have drawn inspiration from representations typically used for natural
images, and have shown that popular natural image representations such as the \emph{bag-of-visual-words} (BoV) \cite{Csurka2004} or the \textit{Fisher-Vector}
\cite{Perronnin07} built on top of densely-extracted local descriptors such as SIFT \cite{lowe04ijcv} or SURF \cite{herbert06} lead to notable improvements.  All the latter representations are in general
task-agnostic. They get combined with the right algorithm, such as a classifier, or a clustering method, in order to produce the right prediction depending on
the target application. These shallow features were shown to generalize very well across tasks~\cite{docimgArXiv16}.

Recently, \textit{deep features}, and convolutional neural networks (CNN) in particular,
were  applied to document images and have shown better classification  and retrieval performances than 
some shallow features (BoV)~\cite{harleyetal15, lekang14}.
  The characteristic of deep features is that they are learnt
\textit{end-to-end}. This means that the two previously distinct steps of i) feature
construction and ii) prediction (classification in most of the cases) are
merged into one step. In other words, the feature and the classifier are learnt
jointly and cannot be distinguished any more. They have been recently shown to outperform
some shallow features (BoV) by a large margin \cite{harleyetal15},  but they are highly specialized for a specific task, and their use as a generic feature extractor for document images has not been studied in detail. 
Also, they are a lot more costly to train, as learning can easily take several days on a GPU.

Hybrid architectures were recently introduced \cite{perronninlarlus15} to classify natural images. They also showed good transferability properties from 
classification to the image retrieval task. We are not aware of any work where these hybrid models have been applied to document images.

\section{Feature Representations for Document Images}
\label{sec:representation}

In this paper, we consider a broad range of feature representations for document images.  First, we select two shallow 
features that were successfully used in
various document image tasks~\cite{GordoICDAR13,Gordo13,docimgArXiv16}: the Runlength feature~\cite{ChanChang01} (Section \ref{sec:rl}) and 
the Fisher-Vector~\cite{Perronnin07} representation (Section \ref{sec:fv}). 
We also experiment with deep features, more precisely two different convolutional neural network architectures the 
AlexNet~\cite{krizhevsky2012cnn} and GoogLeNet~\cite{Szegedy2015}  described in Section \ref{sec:cnn}. 
Finally, we  briefly recall from~\cite{perronninlarlus15} the hybrid architecture in Section \ref{sec:hybrid}, 
that is the first time used for document image representation.

\subsection{\bf RunLength features}
\label{sec:rl}

\begin{figure}[t]
\centering
\includegraphics[width=0.3\linewidth]{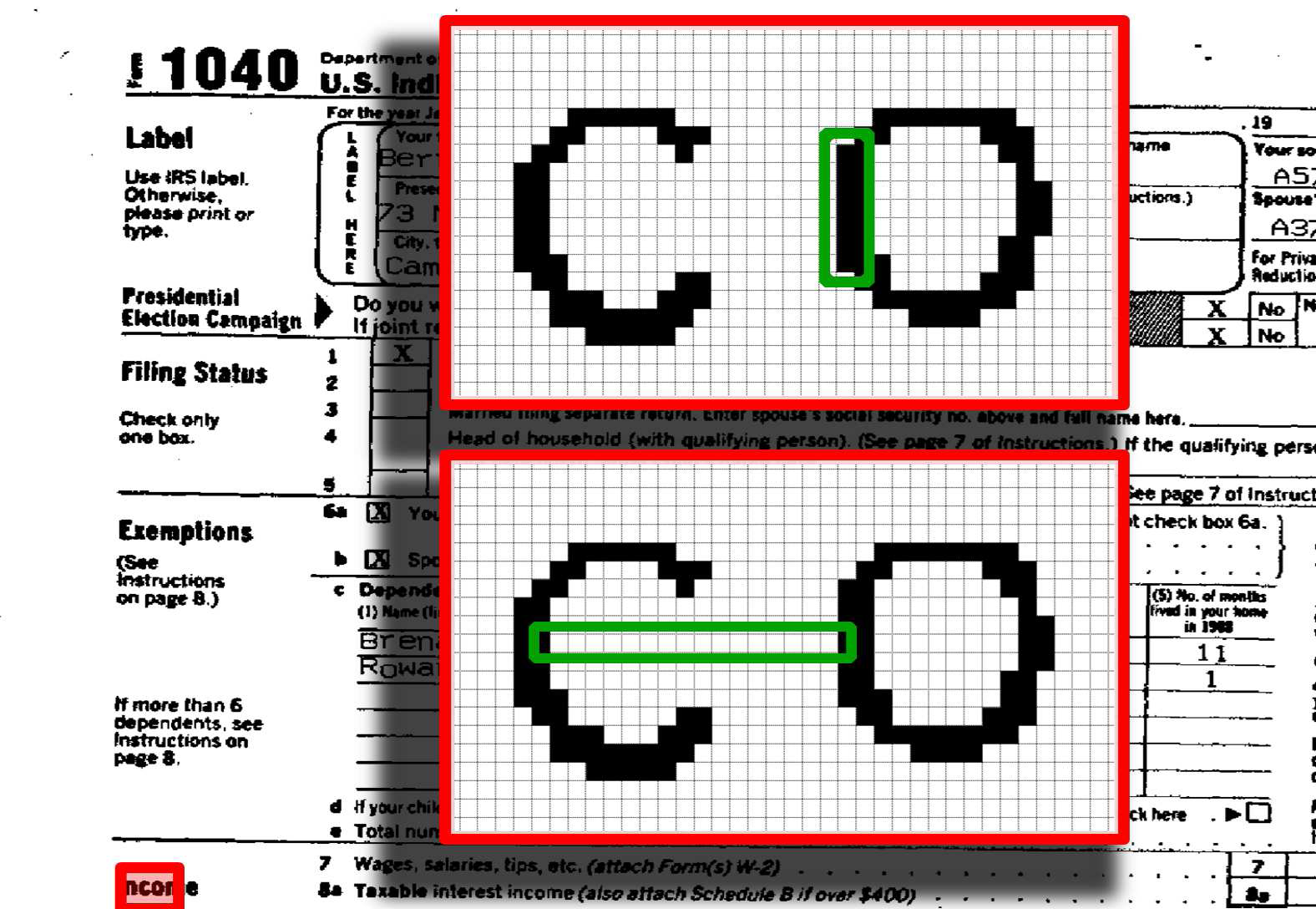}\hspace{1cm}
\includegraphics[width=0.6\linewidth]{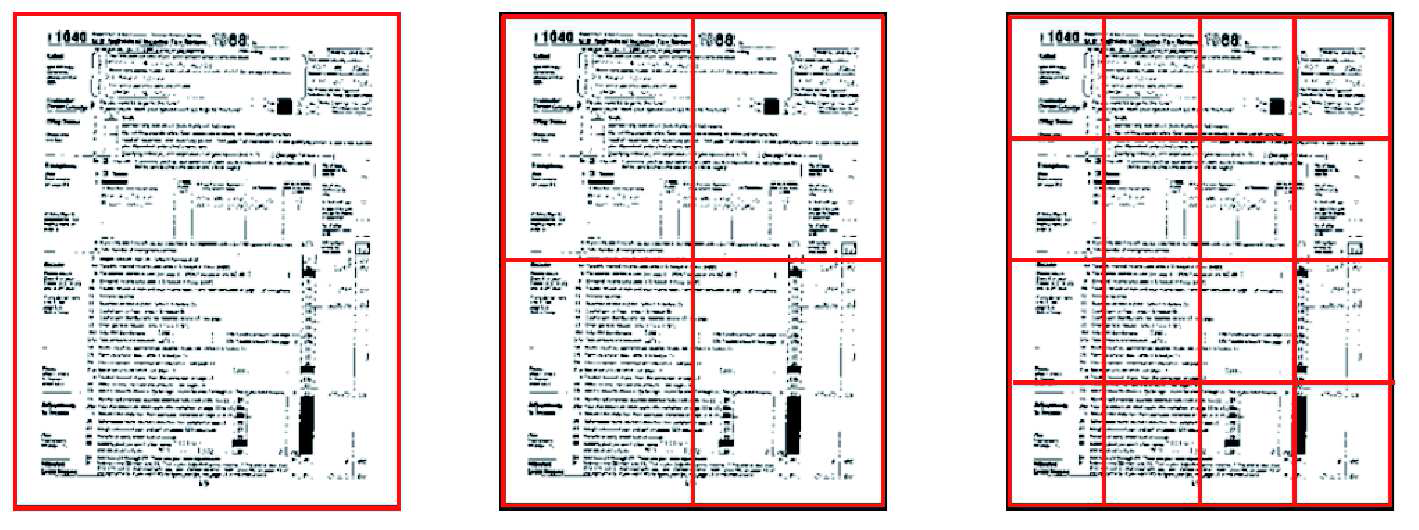}
\caption{Illustration of the Runlength (RL) feature. Left: 2 pixel runs (a
  vertical black run of length 7, shown on the top image and
a horizontal white run of length 16, shown in the bottom)
are extracted from a small region on the bottom-left corner of the document
(shown in red). 
Right: A tree layer spatial pyramid captures the document layout (image courtesy of \cite{Gordo13}).}
\label{fig:RLex}
\end{figure}

The main intuition behind the RunLength (RL) features \cite{ChanChang01} is to encode sequences of pixels
that share the same value and that are aligned (\eg vertically, horizontally
or diagonally). The "run-length" is the length of those sequences (see \eg the
green rectangles in the Figure~\ref{fig:RLex}). 

While the RL feature can be
extended to consider sequences of similar gray-scale or even color values, 
binary images are sufficient in practice to characterize document
images~\cite{Gordo13,GPV13}. Therefore, we first binarize the
document images and consider only runs of black and white pixels. In case of
color images, we binarize the luminance channel using a simple thresholding at
0.5 (where image pixels intensities are represented between 0 and 1). More
complex binarization techniques exist (see \eg participations in the DIBCO and
HDIBCO~\cite{Pratikakis12} contests), however testing them is out of the scope of this paper. 

Note that optionally,
we can resize the images after binarization to have the same resolution within
the dataset. In our experiments, we select a maximum number of pixels (250M) and 
we downscale all images that are larger, keeping the aspect ratio, but we do not upscale 
images that are below this target size.

On the binarized images, the numbers of (black or white) pixel runs are
collected into histograms. As suggested in \cite{Gordo13,GPV13}, we use a logarithmic quantization of the lengths to
build these histograms in order to be less sensitive to noises and small
variations :
\begin{equation*}
 [1], [2], [3-4], [5-8], [9-16] ,  \ldots, [\geq (2^q+1)].
\end{equation*}

This yields two histograms of length $Q=q+2$ per direction, 
one for the white pixels and one for the black pixels. We compute these runs in four directions, 
horizontal, vertical, diagonal and anti-diagonal,  
and concatenate all the obtained histograms. An image (or image region) is then represented by this  
$4\times2\times Q$ dimensional RL histogram.

In order to better capture information about the page layout we use a spatial
pyramid~\cite{LSP06} with several layers such that at each level the image is
divided into $n \times n$ regions and the RL histograms computed on these regions
are concatenated to obtain the full image signature (see illustration in
Figure~\ref{fig:RLex}). 
To obtain the final RL image feature, we L1-normalize and
apply component-wise squarooting as in \cite{GPV13}. 
As in \cite{docimgArXiv16} best performances were obtained with 
5 Layers ($1 \times 1,2\times 2,4\times 4,6\times 6,8\times 8$) and $Q=11$, we use this configuration and hence
 in our experiments the final RL features are of $121*8*11=10648$ dimensional.

\begin{figure}
\centering
\includegraphics[width=\linewidth]{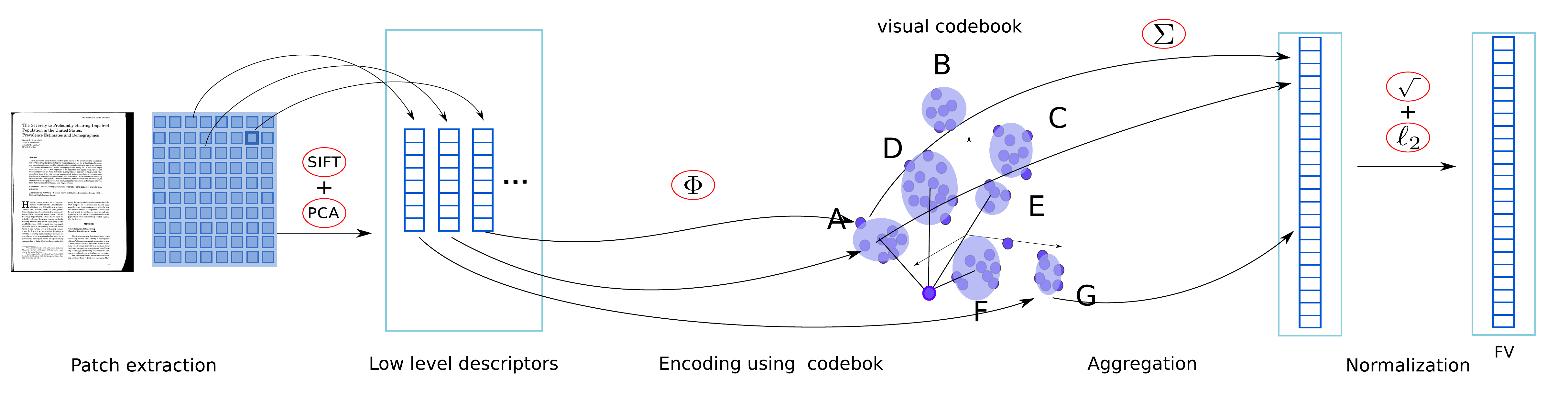}
\caption{Illustration of the Fisher-Vector representation \cite{Perronnin07}.}
\label{fig:fv}
\end{figure}

\subsection{\bf Fisher-Vector representations}
\label{sec:fv}

The Fisher-Vector (FV)~\cite{Perronnin07} can be seen as an extension of the
bag-of-visual-words (BoV)~\cite{Sivic03,Csurka2004} that
goes beyond simple counting (0-order statistics) and that encodes higher order
statistics about the distribution of local descriptors assigned to visual words.
Similarly to the BoV, the FV depends on an intermediate representation: the
visual vocabulary. The visual vocabulary can be seen as a probability density
function (pdf) which models the emission of the low-level descriptors in the
image. We represent this density by a Gaussian mixture model (GMM).

The FV characterizes the set of low-level features (in our case SIFT features~\cite{lowe04ijcv}),
 $X_I=\{\bm x_t\}_{t=1}^{T}$ extracted from an image $I$ by encoding necessary
modifications of the GMM model in order to best fit this particular
feature set. Assuming independence, this can be written as the gradient of the
log-likelihood of the data on the model: 
\begin{equation}
G_{\lambda}(I)  = \frac{1}{T} \sum_{t=1}^{T} \nabla_{\lambda}
\log \left\{\sum_{n=1}^N w_n {\mathcal N}(\bm x_t | \mu_n,\Sigma_n)\right\}
\end{equation}
where $w_n$, $\mu_n$ and $\Sigma_n$ denote
respectively the weight, mean vector and covariance matrix of the Gaussian $n$
and $N$ is the number of Gaussians in the mixture. 

To compare 
two images $I$ and $J$, a natural kernel on these gradients is  the Fisher Kernel
$K(I,J) = {G_{\lambda}(I)}^{\top} F_{\lambda}^{-1} G_{\lambda}(J)$, 
where $F_{\lambda}$ is the  Fisher Information Matrix. As $F_{\lambda}^{-1}$  
is symmetric and positive definite, it has a Cholesky decomposition
$L_{\lambda}^{\top} L_{\lambda}$ and  $K(I,J)$ can be rewritten as a 
dot-product between normalized vectors $\Gamma_{\lambda}$ where: 
\begin{equation}
\Gamma_{\lambda}(I) =  L_{\lambda} G_{\lambda}(I)
\end{equation} 
to which we refer as the {\em Fisher-Vector} (FV) of the image $I$. 

Following \cite{Perronnin07,PSM10}, we assume diagonal covariance matrices in
the GMM and ignore the gradients with respect to the weights.
We obtain the following normalized gradients :
\begin{eqnarray}
\Gamma_{\mu^d_n}(I)  & = & \frac{1}{T\sqrt{w_n}} \sum_{t=1}^T 
g_n(\bm x_t) \left( \frac{\bm x^d_{t}-\mu^d_{n}}{s^d_{n}} \right) \label{eqn:dm} , \\
\Gamma_{s^d_n}(I) &  = & \frac{1}{T\sqrt{2 w_n}} \sum_{t=1}^T 
g_n(\bm x_t) \left[ \frac{(\bm x^d_t-\mu^d_n)^2}{(s^d_n)^2} -1 \right] 
\label{eqn:ds} 
\end{eqnarray}
where $g_n(\bm x_t)=\frac{w_n {\mathcal N}(\bm x_t | \mu_n,\Sigma_n)}
{\sum_{j=1}^N w_j {\mathcal N}(\bm x_t | \mu_j,\Sigma_j)}$ and
$s^d_{n}$ are the elements of the  diagonal $\Sigma_n$.
The final gradient vector $\Gamma_{\lambda}(I)$ concatenates all
$\Gamma_{\mu^d_n}(I)$ and $\Gamma_{s^d_n}(I)$, and is $2ND$-dimensional, where $D$
is the dimension of the low level features $\bm x_t$.
As proposed in \cite{PSM10} we apply a component-wise squarooting
followed by L2-normalization to produce the final Fisher-Vectors. The full process is illustrated in Figure~\ref{fig:fv}.

In our experiments we consider either this image-level FV with a large number of Gaussians in the vocabulary 
($N=256$) or a spatial pyramid version\footnote{We use a single-layer for the spatial-pyramid. Initial
  experiments with multiple layer spatial pyramids as in the case of RL did not improve results.}
  with smaller vocabulary sizes (pyramid $4\times 4$ is combined with $N=16$ and $8\times 8$ with $N=4$).  
  This consistently yields a 40960 dimensional vector
representation\footnote{We reduce SIFT features from 128 to 77 dimension, and add the center and the 
scale of the patch in order to capture some location information, \ie $D=80$.}.

\subsection{\bf Convolutional Neural Networks}
\label{sec:cnn}

Convolutional Neural Networks (CNNs) are composed of several layers that combine linear as well as non-linear operators jointly learned,
 in an end-to-end manner, to solve a particular task. Typically, they have a standard structure: stacked convolutional layers (optionally combined with contrast normalization and max pooling),
followed by one or more fully-connected layers, and a softmax classifier as the final layer. 
Therefore, a feed-forward neural network can be thought of as the
composition of a number of functions 
\begin{eqnarray}
F(x) = F_L(...F_2(F_1(x, W_1), W_2), ..., W_L),
\end{eqnarray}
where each function $F_i$ takes as input $x_i$ and a set of
parameters $W_i$, and produces $x_{i+1}$ as output.

Convolutional layers are the core building blocks of CNNs and consist of a set of small and \emph{learnable} filters 
that extend through the full depth of the input volume and slides across width and height. 
Max pooling layers are inserted in-between successive convolutional layers in order to progressively 
reduce the spatial size of the representation and the amount of parameters of the network. 
Hence they also control the over-fitting.  Local contrast operation layers are used to normalize the responses across feature maps. 
The fully-connected layers are linear projections, \ie matrix multiplications 
followed by a bias offset, where the neurons are connected to all activations of the previous layer.
CNNs also use \emph{ReLU} non-linearities ($relu(x) = \max(0,x)$), which rectify the feature maps to 
ensure they remain positive. 

Although the architecture of these networks, which is defined by the hyperparameters and the arrangement of these blocks, 
are commonly handcrafted, the parameters set $W_1,\ldots,W_L$ of the network are learned 
in a supervised manner from a set of $M$ labeled images $\{I_m,y_m\}$ using a suitable loss function for the task at hand.  These
are trained by back-propagation using stochastic gradient descent (see details of the training procedure in Section \ref{sec:training}).

Since their introduction in the early 1990's (LeNet) \cite{LeCun1989}, and mostly since their recent success in various challenges including the ImageNet Large
Scale Visual Recognition Challenge (ILSVRC)~\cite{russakovsky2014imagenet}, many different CNN architectures have been proposed
\cite{krizhevsky2012cnn,Szegedy2015,zeiler2014,simonyan2014verydeep}. In this paper we focus on two popular ones: AlexNet \cite{krizhevsky2012cnn} and GoogLeNet
\cite{Szegedy2015}.

\begin{figure}[t]
\centering
\includegraphics[width=0.9\linewidth]{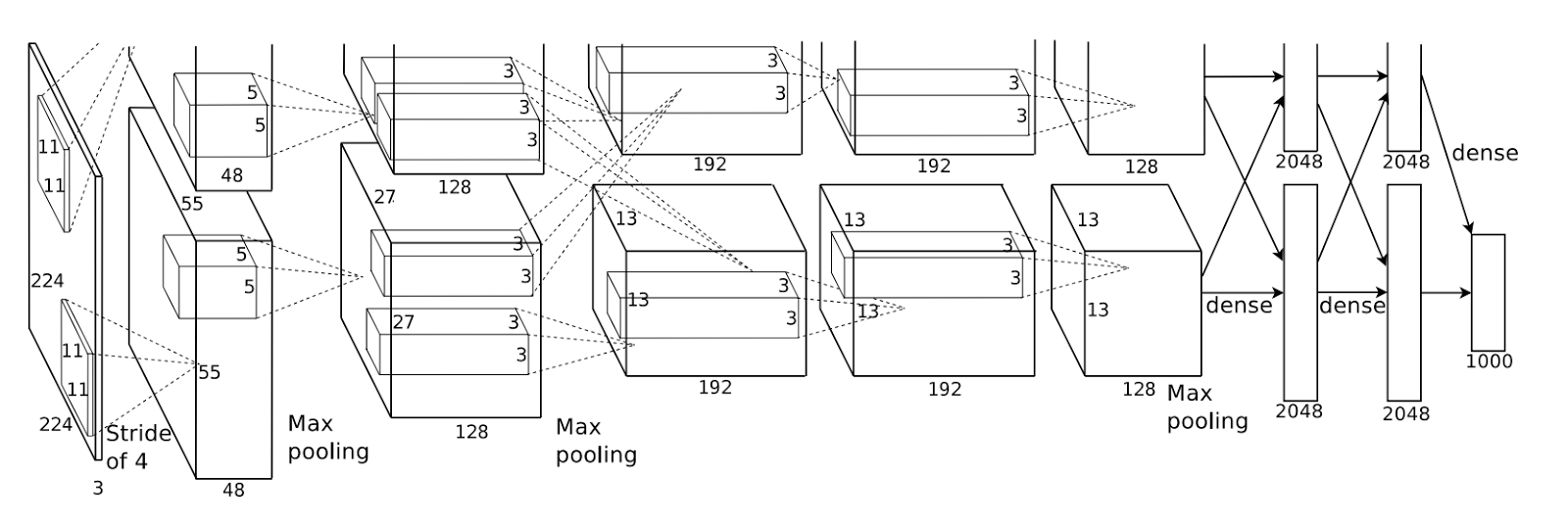}
\caption{The AlexNet architecture (image courtesy of \cite{krizhevsky2012cnn})}
\label{fig:alexnet}
\end{figure}

\myparagraph{AlexNet}
The AlexNet architecture, proposed by Krizhevsky \etal~\cite{krizhevsky2012cnn}, was the first successful CNN architecture for the image classification task,
outperforming by a large margin shallow methods in the ILSVRC 2012 competition. This network is composed of eight layers with weights. Five convolutional layers
with (96, 256, 384, 384, 256) kernels of sizes (11, 5, 3, 3, 3) and a stride of 1 pixel, except for the first layer that has a stride of 4. A response
normalization is applied after layers 1 and 2, and a max pooling with size 2 and a stride of 2 pixels are applied after layers 1, 2, and 5. 
This is followed by
three fully connected layers of sizes 4096, 4096, and $C$ respectively, where $C$ is the number of classes. The output of the last fully-connected layer is
fed to a $C$-way softmax which produces a distribution over the $C$ class labels. A ReLU non-linearity is applied after every convolutional or fully connected
layer. The network is fed with fixed-size $224 \times 224 \times 3$ images\footnote{Note that in contrast to shallow features where the aspect ratio is kept 
when resizing the images, here the aspect ratio can be modified.}. The architecture is summarized in Figure \ref{fig:alexnet}.

\myparagraph{GoogLeNet} Szegedy \etal set a new state-of-the-art in image classification and object recognition in the ILSVRC 2014 competition with a
significantly different architecture, the GoogLeNet \cite{Szegedy2015}. It uses a deeper and wider architecture than 
traditional CNNs, with 10 times fewer parameters compared to standard CNNs. 

\begin{figure}[t]
\centering
\includegraphics[width=0.6\linewidth]{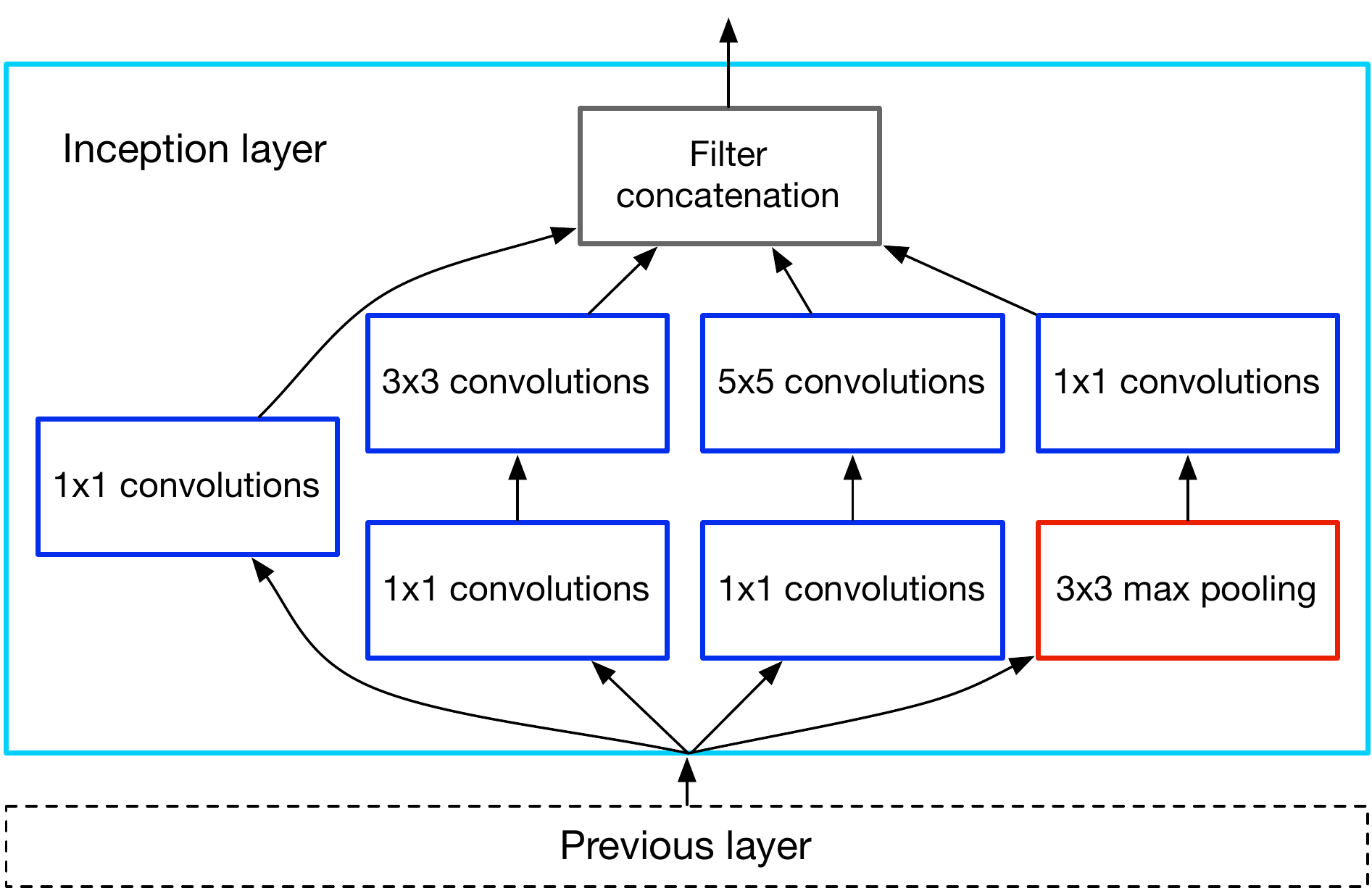}
\caption{Inception module as used in the GoogLeNet architecture \cite{Szegedy2015}.}
\label{fig:inception}
\end{figure}

The main idea behind  is the {\em inception architecture}, 
based on finding out how an optimal local sparse structure in a convolutional neural network can
be approximated and covered by readily available dense components.  For that, GoogLeNet relies on several {\em inception layers} 
(Figure~\ref{fig:inception}), where each such layer uses a series of trainable filters with sizes 
$1 \times 1$, $3 \times 3$ and $5 \times 5$. In this way, there are multiple filter sizes per layer, so each layer has
the ability to target the different feature resolutions that may occur in its input. 
In order to avoid computational blow up, it also performs dimensionality
reduction by  $1 \times 1$ convolutions inserted before the expensive $3 \times 3$ and $5 \times 5$ convolutions. Finally, 
the inception module also includes a parallel pooling path, which is concatenated along with the output of the convolutional 
 layers into a single output vector forming the final output.

One benefit of the inception architecture is that it allows for increasing the number of units at each stage significantly 
without an uncontrolled blow-up of the  computational complexity. Therefore, an inception-based network is a network 
consisting of modules of the above type stacked upon each other, with occasional
max-pooling layers with stride 2 to halve the resolution of the grid. Another important characteristic of 
this architecture is that it uses average pooling instead of
fully connected layers at the top of the last inception layer, eliminating in this way a large amount of parameters.

\begin{figure}
\centering
\includegraphics[width=0.9\linewidth]{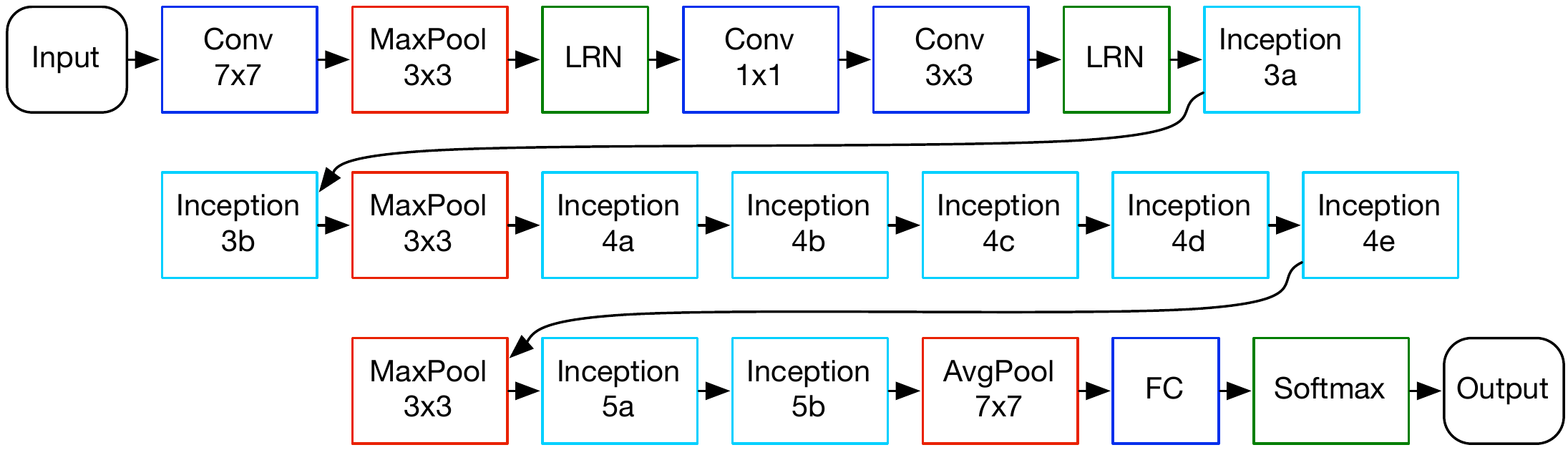}
\caption{Overview of the GoogLeNet architecture \cite{Szegedy2015}.}
\label{fig:googlenet}
\end{figure}

The GoogLeNet architecture that won the ILSVRC2014 challenge is shown in Figure~\ref{fig:googlenet}. It is a network of 22 layers
where  nine inception modules  are stacked after two  convolutional layers with filter sizes of 7 and 3 and strides of 2 and 1. 
Max pooling layers with size 3 and stride 2 are inserted after convolutional layers 1 and 2, and after the inception layers 3b, 4e, and 5b.
An average pooling layer with size 7 and stride 1 follows the last inception layer 5b,  
whose output is fed to a single fully-connected layer and the $C$-way softmax classifier. 
All the convolutions, including those inside the inception modules, use rectified linear activation (\emph{ReLU}).

\subsection{\bf Hybrid descriptors}
\label{sec:hybrid}

\begin{figure}
\centering
\includegraphics[width=\linewidth]{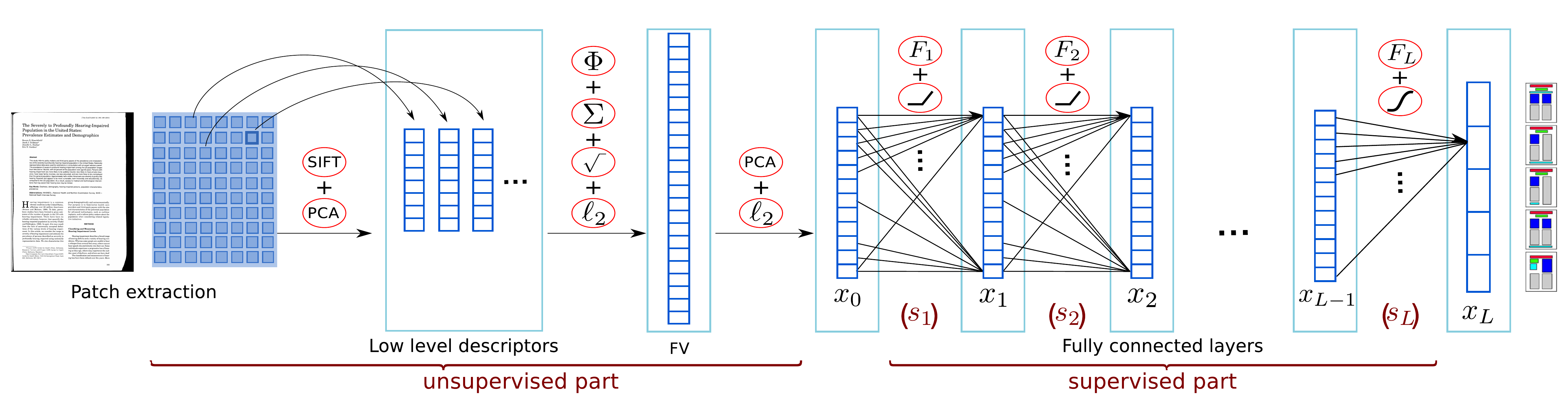}
\caption{Illustration of the hybrid architecture~\cite{perronninlarlus15}.}
\label{fig:hybrid}
\end{figure}

Our last representation is a hybrid descriptor~\cite{perronninlarlus15} that is built using a hybrid architecture 
drawing inspiration from both FVs and CNNs. This architecture combines an
unsupervised part, obtained by an image-level patch-based Fisher-Vector encoding, and a
supervised part composed of fully connected layers.
The intuition of this model is to replace the convolutional layers of the CNN
architecture with a FV representation and to learn subsequent fully-connected
layers in a supervised way, akin to a Multi-Layer Perceptron (MLP), trained with
back-propagation. We provide details below for the  resulting hybrid architecture, illustrated in
Figure~\ref{fig:hybrid}.

The \textit{unsupervised part} of the hybrid architecture is identical to the FV representation
described in section~\ref{sec:fv}. We consider two versions. In the first one, the FV representation is
followed by a PCA projection and L2-normalization, as originally proposed in
\cite{perronninlarlus15}. Alternatively, we consider a hybrid architecture that directly
builds on the full dimensionality FVs, in which case, the dimensional reduction is
performed implicitly by the first fully connected layer of the supervised part
of the architecture (this would be illustrated by a modified Figure \ref{fig:hybrid} where $x_0$ = FV directly).

The \textit{supervised part} uses a set of $L-1$ fully connected layers $x_1, \ldots x_{L-1}$  of sizes 4096
and a last layer of size $C$, where $C$ is the number of classes, and a ReLU
non-linearity is applied after every fully connected layer. 
Like in AlexNet and GoogLeNet, the output of the last fully-connected layer is fed to a $C$-way
softmax which produces a distribution over the $C$ class labels.

As an alternative, we also replace the unsupervised part of this architecture by RunLength histograms described in section~\ref{sec:rl}.

\section{Training}
\label{sec:training}
As shallow, deep and hybrid features require different learning paradigms, this section details the training procedure for them
in the context of the two scenarios that we consider in the
experimental part. In the first one, training and testing are done on the same dataset, 
for the same task (Section~\ref{sec:training1}). In the second one, both
the dataset and the task can vary, and a transfer mechanism is needed (Section~\ref{sec:training2}).

\subsection{\bf Training for the task at end}
\label{sec:training1}

The RL does not require any training, all the parameters are already predefined, so this descriptor is truly dataset-agnostic. 
The FV requires a visual codebook that is learned in an unsupervised manner (by clustering local features extracted from the training set). 
Beyond this unsupervised training step, this descriptor
does not depend on the data, and more importantly on the labels, hence it is independent from the task. 
To solve a classification problem, document images features
and document labels are used to train a classifier. In all our experiments we use a linear 
Support Vector Machine (SVM) classifier~\cite{Vapnik1998} 
on top of RL or FV features.

Unlike the previous two representations (RL and FV), CNNs are deep learning approaches that group feature extraction and prediction into a single
architecture. Consequently, features are learned to optimize the prediction task, and the classifier is already integrated in the architecture. 
Therefore, at test time,  the full architecture is used to predict the document label.

The set of parameters $W$ of the CNN, which includes filters in the convolutional layers, weight matrices and biases in the fully-connected layers, are learned in
a supervised manner from a set of $N$ labeled images $\{I_n,y_n\}$ using a suitable loss function for the task to be solved. In our case, we will train both
AlexNet and GoogLeNet for document image classification.
To train the parameters, we use the standard objective that involves minimizing the cross-entropy between the network output and the ground-truth:
\begin{equation}
\sum_{n=1}^N \sum_{c=1}^C y_{n,c} \log(\hat{y}_{n,c})
\end{equation}
where $y_{n,c}$ is the ground-truth label, $\hat{y}_{n,c}$ is the prediction of label $c$ for image $n$ as output by the last layer, $C$ is the number of
classes and $N$ is the number of available training examples.
We update the parameters via stochastic gradient descent (SGD)~\cite{bottou10compstat} by back-propagating the derivative of the loss with respect to the
parameters throughout the network. To avoid over-fitting, we use drop-out~\cite{krizhevsky2012cnn} at the input of the fully-connected layers.

The hybrid descriptors share similarities both with traditional features 
(FV or RL depending on what is used in the unsupervised part) and with deep
features. In the case of FV, the unsupervised part requires learning the visual codebook on patches extracted from the dataset. 
The supervised part is trained end-to-end  with the classifier integrated in the architecture. 
To learn the parameters of the supervised part, as for  the CNNs, we
minimize the cross-entropy between the label predicted by the last layer and 
the ground-truth labels. Weights are updated using back-propagation. Again, we use
drop-out.

\subsection{\bf Training for a different task}
\label{sec:training2}

In the case of shallow features (RL or FV), descriptors can be used for a different task, and they only need to be combined 
with the right predictor (ranking, new classifier, clustering algorithm, etc.).

In the case of CNNs, besides its common use to solve a given task in an end-to-end manner, it has also become a standard practice to use them as feature extractors. 
Convolutional filters in the first layers can be seen as detectors of basic structures, like corners or straight lines, while deeper layers are able to capture more complex structures and semantic information. 
Therefore, a given image can be feed-forwarded through the CNN and the activations of intermediate layers  used as mid-level features to
represent it. These off-the-shelf features can be subsequently combined with the right prediction algorithm. 
This finding was quantitatively validated for a number of tasks including image classification~\cite{donahue2014decaf,oquab2014transferring,zeiler2014,chatfield2014return,Razavian2014}, image
retrieval~\cite{Razavian2014,Babenko2014}, object detection~\cite{girshick2014rich}, and action recognition~\cite{oquab2014transferring}. 
We show that these findings also generalize for document images.

For the CNNs, we extract features from different layers at different depths and compare their performance in the experimental section. 
In the case of AlexNet, we use as features the output activations of the last 
convolutional layer (pool5), and the output of the 
first two fully-connected layers (fc6 and fc7). 
In the case of GoogLeNet, we consider the output of different inception layers and the output of the average 
pooling layer previous to the fully-connected layer
(p5s1). Concretely, we use inception layers 3a, 3b, 4a, 4e and 5b (see Figure~\ref{fig:googlenet}). 
For the hybrid architecture, we  experimented with  the output activations of different fully-connected layers.  
We L2 normalize the activation features from both CNNs and hybrid architectures 
before feeding them in the desired predictor.

\begin{table}[t]
\small
\begin{center}
\begin{tabular}{|c|c|c|c|c|}
\hline
dataset & \# images &  image size & \# categ. & category description\\
\hline
RVL-CDIP & 400000 & 750M & 16  & document types \\
NIST & 5590  & 8.4M  & 12   & form types\\ 
MARG & 1553 & 8.4M & 9 & layout types  \\
CLEF-IP  & 38081 & 1.5K - 4.5M & 9 &  patent image types\\
IH1 & 11252 &  1.2M  & 14 &   document types \\
IH2 & 884 & 1.4M & 72 &   document types and layout \\
IH3 & 7716  & 0.5M - 5M &   63 & fine-grained document types \\
\hline
\end{tabular}
\caption{Statistics of the seven datasets considered in our experiments.}
\label{tab:statistics}
\end{center}
\end{table}

\section{Evaluation framework}
\label{sec:eval}

In this section, first, we describe the datasets (Section \ref{sec:dataset}) and then 
 we provide implementation details concerning our experiments (Section \ref{sec:framework}).

\subsection{\bf Datasets}
\label{sec:dataset}

\begin{figure}[t]
\centering
\includegraphics[width=\textwidth]{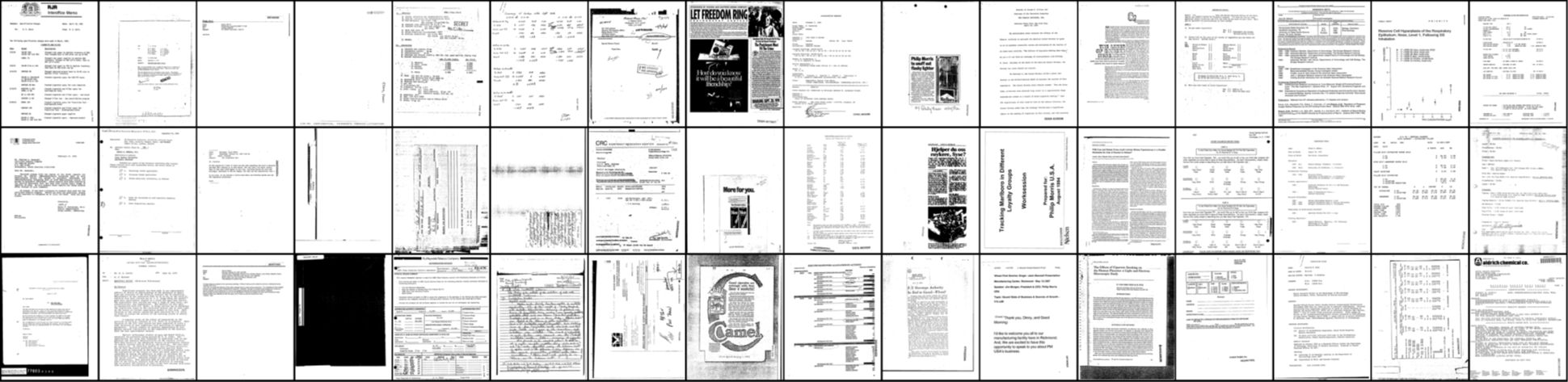}
\caption{Illustration of the RVL-CDIP dataset. Each column corresponds to one of the 16 classes
(image courtesy of \cite{harleyetal15}).}
\label{fig:tobacco}
\end{figure}

\begin{figure}[t]
\centering
\includegraphics[width=\textwidth]{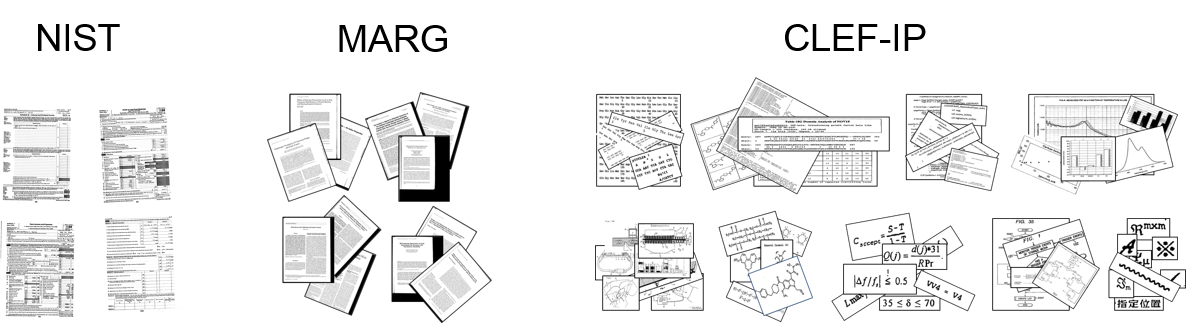}
\caption{Example images from the NIST, MARG and
CLEF-IP datasets.}
\label{fig:benchmarkdataex}
\end{figure}

\begin{figure}[t]
\centering
\includegraphics[width=\textwidth]{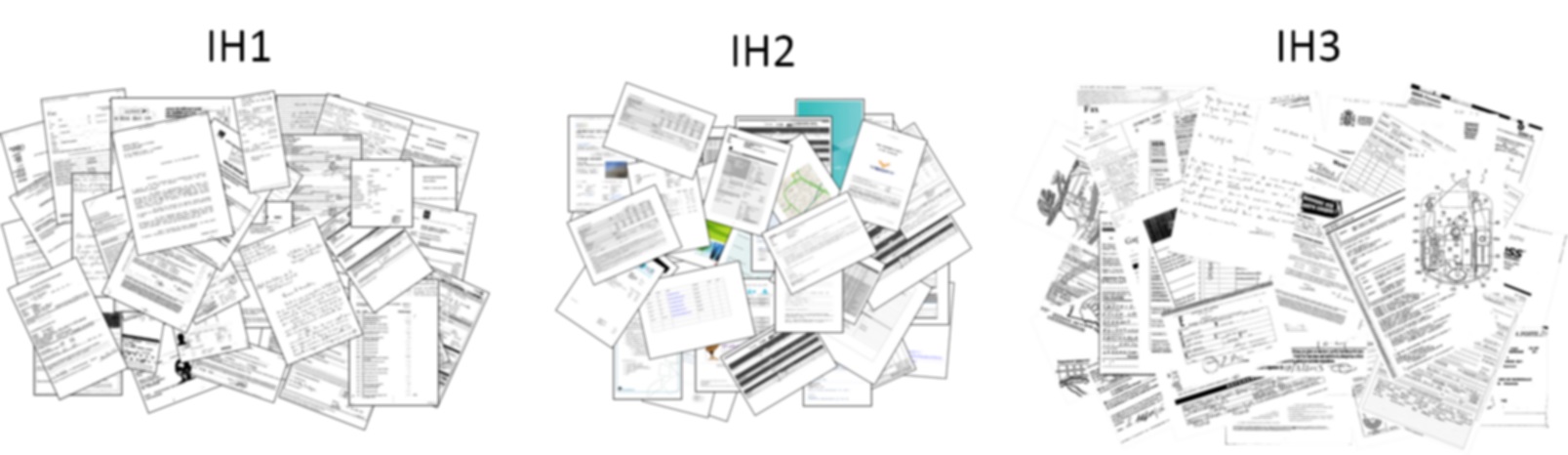}
\caption{Example images from the 3 in-house customer dataset IH1, IH2 and IH3. The images were intentionally blurred for privacy reasons. 
Here we show examples to illustrate the variability of documents within each dataset.}
\label{fig:inhouseex}
\end{figure}

We conducted a broad experimental study comparing the feature representations
described in the previous section on seven different datasets.
We used four publicly available datasets, namely RVL-CDIP, NIST, MARG, and
CLEF-IP. We also confirm our conclusions on three in-house customer datasets,
that we refer to as IH1, IH2, and IH3. Statistics on the different datasets
can be found in the Table \ref{tab:statistics} and some illustrations in 
Figures~\ref{fig:tobacco}, \ref{fig:benchmarkdataex} and \ref{fig:inhouseex}. 
We detail their characteristics below.

\paragraph{\bf RVL-CDIP} The Ryerson Vision Lab Complex Document Information
Processing (RVL-CDIP) dataset\footnote{\url{http://scs.ryerson.ca/~aharley/rvl-cdip/}}~\cite{harleyetal15} is a
subset of the IIT-CDIP Test collection~\cite{Lewis2006}. It is composed of 400000 images 
labeled with one of the following 16 categories: {\em letter, memo,
email, filefolder, form, handwritten, invoice, advertisement, budget, news article, presentation, scientific publication, questionnaire, resume, scientific
report}, and {\em specification}. Figure~\ref{fig:tobacco} shows 3 examples of each class.
  
\paragraph{\bf NIST} The NIST Structured Forms Reference Set\footnote{\url{http://www.nist.gov/srd/nistsd2.cfm}}~\cite{nistforms91} is a dataset
of black-and-white images that consists of 5590 pages of synthesized documents.  These documents correspond to 12 different 
tax forms from the IRS 1040 Package X for the
year 1988 (see examples in Figure~\ref{fig:benchmarkdataex}).  Class names are \textit{Forms 1040, 2106, 2441, 4562, 6251} and
\textit{Schedules A, B, C, D, E, F, SE}.

\paragraph{\bf MARG} The Medical Article Records Ground-truth (MARG) 
dataset\footnote{\url{https://ceb.nlm.nih.gov/proj/marg/marg.php}}~\cite{marg03}
consists of 1553 documents, each document corresponding to the first page of
a medical journal.  The dataset is divided into 9 different layout types. These
layouts vary in relative position of the title, the authors, the affiliation,
the abstract and the text (see examples from four classes in Figure~\ref{fig:benchmarkdataex}).
Within each layout type, the document can be
composed of one, two or three columns. This impacts visual similarity a lot and
makes classification and even more clustering on this dataset very challenging.

\paragraph{\bf CLEF-IP}
The CLEF-IP dataset is the training set\footnote{\url{http://www.ifs.tuwien.ac.at/~clef-ip/download/2011/index.shtml}}
released for the Patent Image Classification task of the Clef-IP 2011
Challenge~\cite{Piroi11}.  In the challenge, the aim was to categorize patent
images (\ie figures) into 9 categories: \textit{abstract drawing, graph, flowchart, gene sequence, program listing, symbol, chemical structure, table}
and \textit{mathematics}. We show example images grouped by class in Figure~\ref{fig:benchmarkdataex}. The dataset contains between 300 and 
6000 labeled images for each class, 38081 images in total, with a large variation of the 
image size (from as little as 1500 pixels to more than 4M pixels) and aspect ratio (from 1 to more than 10).

\paragraph{\bf IH1} The first in-house dataset (IH1) regroups internal document
images from a single customer. It contains 11252 scanned documents from 14
different document categories such as \textit{invoices, contracts, IDs,
coupons, handwritten letters,} etc. 

\paragraph{\bf IH2} The second in-house dataset (IH2) is a small dataset of 884
multi-page documents\footnote{We only consider the first page.} from a single customer, 
divided into 72 fine-grained categories representing both the
document type such as \textit{invoice, mail, table, map} and the document layout
(\eg ``\textit{a mail with an excel table on the bottom}'', ``\textit{a table with black lines
separating the rows}'').

\paragraph{\bf IH3} The third in-house dataset (IH3) contains 7716 documents collected from several customers. 
We divided the dataset into 63 fine-grained categories such as diverse types of \textit{forms, invoices, contracts},
etc,  where the class labels were defined on one hand by the generic document type but also by their origin. 
Hence invoices, mails, handwritten or typed letter that belongs to different customers were considered as 
independent classes. The aim with this dataset was to go beyond generic 
document types or layout and simulate real document image based applications where 
documents from several customers should be processed all together (\eg in a print or scan flow).

\subsection{\bf Implementation details}
\label{sec:framework}
\vspace{0.3cm}

\noindent Here we summarize the experimental details of our study.

\paragraph{Shallow features} 
For the Runlength histograms (RL) descriptor, we use a 5-layer pyramid ($1
\times 1,2\times 2,4\times 4,6\times 6,8\times 8$) and 11 quantization levels
(1, 2, 4, \ldots, 512, larger than 512) yielding to  10648 dimensional features.
Images were binarized (when necessary) and rescaled to 250K pixels as suggested in \cite{docimgArXiv16}. 
These features, which are truly dataset-independent, we use in both experimental parts.

Our Fisher-Vector descriptors are built on top of SIFT features extracted at 5 different scales 
(the patch size varies from $24\times24$ to $96\times96$ in
images rescaled to 250K pixels).  The original SIFT features are projected using PCA to a 77-dimensional vector to which we concatenate the position (x,y) and
the scale (s) of the patch, obtaining 80-dimensional local features. We consider different visual vocabulary sizes (4, 16 and 256 Gaussians in the mixture). For a
fair comparison, the grid of the spatial pyramid varies in order to build FVs of the same dimension (40960). 
For a vocabulary with 4 Gaussians, we concatenate
FVs on an $8\times8$ grid (denoted by FV4), for a vocabulary with 16 Gaussians, we concatenate FVs on a $4\times4$ grid (denoted by FV16) and for a vocabulary
of size 256 we use the FV build on the whole image (denoted by FV256).  
Given a new dataset, we can either compute new SIFT-PCA and GMM to build the FVs, or reuse
the models (PCA and GMM) unsupervisedly trained on the RVL-CDIP dataset. 
We opted for the second strategy for two reasons: first, preliminary experiments have
shown very similar results, and second, our study aims at testing the transferability of the models to new datasets.

\paragraph{Hybrid features} 
Both the RL and FV were considered in the unsupervised part of the hybrid architecture (see Section \ref{sec:hybrid}). We refer to them as FV+MLP and RL+MLP
respectively.
Note that for each feature type (\eg FV256 or FV16) we need to build a different hybrid model. In addition to the model proposed in \cite{perronninlarlus15}
where the size of the original FV is first reduced with PCA (to 4096 dimensions), we also build hybrid models directly on the FV without PCA reduction.  In this
case we fix the first fully connected layer to a size of 4096 letting the hybrid model learn the dimensionality reduction. 
By default, results reported for our
hybrid models do not include PCA. When we do, this is mentioned explicitly (FV+PCA+MLP). 
In the experiments exploring feature transferability (Section \ref{sec:part2}),
we use the activation features corresponding to various fully connected layers of these models trained on the RVL-CDIP dataset (see details
in Sections \ref{sec:hybrid} and \ref{sec:training2}).

\paragraph{CNNs} We consider two popular CNN architectures that were successfully
used to classify natural images: AlexNet and GoogLeNet (see Section \ref{sec:cnn}) denoted by CNN-A and CNN-G respectively.
For both models, we initialize  the CNN  with the models (available online)
trained on the ImageNet classification challenge dataset~\cite{russakovsky2014imagenet}
(ILSVRC 2012), and fine-tune them on the RVL-CDIP dataset. We also
conducted experiments where the models were directly trained on RVL-CDIP, but the
results were 1-2\% below the fine-tuned version. As above, for the feature transferability  experiments (Part 2)
we considered activation features corresponding to the models fine-tuned  on the RVL-CDIP dataset 
(see details in Sections \ref{sec:cnn} and \ref{sec:training2}). 

\section{Experiments}
\label{sec:exp}

The experiments are divided into two parts.  In the first part, Section \ref{sec:part1}, our  set of experiments are
related to large scale document image classification using the RVL-CDIP dataset.  The second part, Section \ref{sec:part2}, 
 is devoted to our feature transfer experiments, where we explore how transferable different image representations, 
 learned on the RVL-CDIP dataset,   are to new datasets and tasks without any extra learning or fine-tuning of the parameters.

\subsection{\bf Part 1: Classification of documents from the same dataset}
\label{sec:part1}

\vspace{0.3cm}

The first part of our experimental analysis focuses on the document image classification task. 
We benchmark the different feature representations introduced in
Section~\ref{sec:representation} on the RVL-CDIP dataset.
We followed the experimental protocol (train/val/test split and evaluation measure) suggested in~\cite{harleyetal15}. 
We used the validation set to choose both
the classifier's parameters (learning rate, number of iterations) as well as the model's parameters (\eg number of layers, drop-out level, etc.).
First, we compare the best flavor of each descriptor type, to give a clear summary of the results, 
then we show deeper analyses for the different models.

\begin{table}[t]
\small
\begin{center}
\begin{tabular}{|c|cc|cc|cc|}
\hline
Features & RL+SVM & FV+SVM &  RL+MLP & FV+MLP & CNN-A & CNN-G \\
\hline
Top-1 & 75.6 & 85.1  &  84.8 & 89.3 & 90.1 &  \bf{90.7}  \\
\hline
\end{tabular}
\caption{Top-1 accuracy for different descriptors on the RVL-CDIP
  dataset.}
\label{tab:summary}
\end{center}
\end{table}

\subsubsection{Overall comparison}

Table \ref{tab:summary} summarizes top-1 accuracy on the RVL-CDIP
dataset for the best version of each flavor of features that we consider in our benchmark.
We can make the following observations.

First, we notice the good performances of CNN models. Both CNN-A and CNN-G outperform other descriptors.
The CNN-A results are consistent with state-of-the art results on the RVL-CDIP
dataset from \cite{harleyetal15} that reports 89.9\% top-1 for its holistic AlexNet-based CNN.
By using a better CNN architecture (GoogLeNet), we manage
to improve over state-of-the art results and get 90.7\% top-1 accuracy.

Second, the hybrid architecture  based on Fisher-Vectors (FV+MLP) yields to a performance 
that is very close to CNN-A. This is an interesting observation as these
models are much faster to train than the CNNs, and no GPU is required.  More generally, we can observe the strong 
performance gain (+9.2\% for RL and +4.2\% for FV)
that is brought by the hybrid architecture compared to these features used in their shallow version and combined with an SVM classifier.

Last and not surprisingly, these experiments confirm previous observations from~\cite{docimgArXiv16} that
FV features outperform RL features on the document image classification task
(both using SVM and MLP).

\subsubsection{Deeper Analysis}

In this section, we study the parameters of the representations, showing that some of
them play a crucial role in improving classification  accuracy. 
\begin{table}[t]
\small
\begin{center}
\begin{tabular}{|c|ccc|}
\hline
Model & FV4 & FV16 &  FV256 \\
\hline
+SVM &  83.1 & 85.1 & 85.0\\
+MLP &  88.3 & 89.3 & 89.1 \\
\hline
\end{tabular}
\caption{Top-1 accuracy on RVL-CDIP for FV vectors using  different vocabulary sizes, combined
  with a SVM classifier (+SVM), or within the hybrid architecture (+MLP).}
\label{tab:FV}
\end{center}
\end{table}

\paragraph{The vocabulary size for the FV}

In Table~\ref{tab:FV} we compare the different FV representations whose visual
vocabulary varies between 4 and 256 Gaussian in the GMM. For these experiments
the grid structure of the spatial pyramid is adjusted to compare representations
of equal length. These representations are combined either with an SVM classifier or used within the
hybrid architecture (\ie MLPs).
We compare FV4 with a vocabulary of 4 Gaussians
and an $8\times8$ grid, FV16 with a vocabulary of 16 Gaussians and a $4\times4$ grid, and finally
FV256 with 256 Gaussians and no spatial pyramid.
We can see that while FV256 and FV16 are on par when we use SVM, the hybrid
model built on FV16 performs slightly better. Also, we observe that FV4 performs worse
than the other descriptors in both cases, showing that the vocabulary needs to
be expressive enough. Therefore, we do not report further results with FV4.

\begin{table}[t]
\small
\begin{center}
\begin{tabular}{|c|cccc|}
\hline
Model & $H=1$ & $H=2$ & $H=3$ & $H=4$\\
\hline
FV16+MLP & 89.3 & 89.3 & 89.2 & 89.2 \\
FV256+MLP & 88.2 & 89.1 & 89.0 & 89.0 \\
\hline
FV256+PCA+MLP & 88.4 & 88.5 & 88.6 & 88.5 \\
\hline
\end{tabular}
\caption{Top-1 accuracy on RVL-CDIP for different number of hidden layers ($H$) in the hybrid
  architecture, for different FV representations, optionally followed by a PCA (last raw).}
\label{tab:layers}
\end{center}
\end{table}

\paragraph{Number of hidden layers in the hybrid architecture}
We first look at the modified hybrid architecture that we proposed, \ie which does not apply PCA to the FV representations in the unsupervised part.  Table
\ref{tab:layers} compares several hybrid architectures. On top of either FV256 or FV16
we use a varying number $H$ of hidden layers,
building increasingly deep architectures.
We observe that even with a small number of layers, we obtain good
performances. Moreover, even a single layer already achieves better results than the
FV+SVM strategy. 
Note that all hidden layers have their size fixed to 4096 but we 
varied the level of drop out. Best results were obtained in general with a drop out level of 30\% or 40\%.

\paragraph{Influence of PCA in the hybrid architecture}
We modified the original hybrid model of \cite{perronninlarlus15}
to remove the PCA projection and to integrate the dimensionality reduction in
the first fully connected layer of the supervised part of the architecture.
In that case, the input of the fully connected layer is the FV without PCA
reduction. In the last raw of the Table~\ref{tab:layers}, we compare the previous results with the original
model (built on top of PCA reduced FVs), still varying the number of hidden
layers $H$. We observe that except for the single layer case ($H=1$), the proposed
hybrid architecture that does not perform PCA but discriminatively learns a
dimension reduction performs better.

\begin{table}[t]
\small
\begin{center}
\begin{tabular}{|c|cccccc|}
\hline
Retrieval (mAP) & NIST  & MARG & CLEF-IP & IH1 & IH2  & IH3 \\
\hline
CNN-A-p5 & \bf{100} & \bf{38.7} & 40.9 & 75.1 & 78.8  & 67.7\\
CNN-A-fc6 & 99.9 & 37.0 & 38.6 & 72.7 &  76.8 & 63.6 \\
CNN-A-fc7 &  93.6 & 30.5 & 28.9 & 59.1 & 65.7 &  43.8 \\
\hline 
CNN-G-i3a &  \bf{100} & 33.6 & 38.4 & 73.4 & 78.3 &  66.2 \\
CNN-G-i3b &  \bf{100} & 34.1 & 36.9 & 77.1 & 80.5 & 67.0 \\
CNN-G-i4a & \bf{100} & 36.4 & 42.4 & \bf{78.5} & \bf{81.8} & 73.4 \\
CNN-G-i4e & \bf{100} & 36.2 & \bf{43.9} & 78.0 & 80.3 & \bf{73.8} \\
CNN-G-i5b & 99.9 & 35.2 &  39.0 & 75.6 & 79.2 & 67.9 \\
CNN-G-p5s1 & 78.9 & 27.8& 32.2 & 63.0 &  64.6 & 44.7 \\
\hline
\end{tabular}
\caption{Retrieval task: mean Average Precision (mAP) for different CNN
  ``off-the-shelf'' descriptors on different transfer datasets.}
\label{tab:CNNMAP}
\end{center}
\end{table}

\subsection{\bf Part 2: Transfer of features to different datasets and tasks}
\label{sec:part2}

\vspace{0.3cm}

In this section we explore how transferable the models and the related features are to new datasets
and tasks. We target three different tasks: i) retrieval, ii) clustering and
iii) (non parametric) classification.
For all the experiments, we assume that the models generating the features (except for RL that needs no extra model) 
have been trained  on the RVL-CDIP dataset, and we apply them to one of the six remaining
datasets.

For all three tasks, we randomly split the datasets in halves, the first set for
training, and the second set for testing. This is done fives times, and we report
averaged results over the five splits.
To asses the performance for a given split we proceed as follows.

\paragraph{Retrieval} Each test example is considered in turn as query example and 
the documents in the training set are ranked according to their similarity to the query.  
As our features are L2 normalized, we used the dot product as similarity measure for all features.
To asses the retrieval performance, we use mean average precision (mAP). We also
computed precision at 1 (P@1) and at 5 (P@5) by averaging the corresponding precision over
all query examples, but as they exhibited similar behavior, we only report the mAP results.

\paragraph{Clustering}
For each split, we cluster samples from the training set using hierarchical clustering with
centroid-linkage into as many clusters as the number of classes we have in the dataset. To evaluate the quality of
the clustering we consider three different measures: the adjusted mutual
information~\cite{VinhEppsBailey09} between true class labels and cluster labels, ii)
the adjusted Random Index~\cite{HubertArabie85}, and iii) the
V-measure~\cite{RosenbergHirschberg07} (which is the weighted harmonic mean of
homogeneity and completeness). As we observed similar trends for these three measures,
we only report results with the adjusted mutual information (AMI).
Note that other clustering algorithms and different numbers of clusters could 
have lead to better performances, however here we are not interested on the
clustering algorithm itself, but on comparing the different features in a
similar setting.

\paragraph{Classification}
We consider the Nearest Classification Mean (NCM) classifier~\cite{mensinkpami13} in our classification
experiments as it is a non-parametric classifier.  In the case of the NCM classifier, each class is represented by
the centroid (class mean) of its training examples and a test element is
assigned to the class of the closest centroid. We report overall classification
accuracy (number of correctly classified test documents divided by the
number of test documents). We could have considered $k$-NN classifiers instead, however as 
the retrieval accuracy P@1 is equivalent to the $k$-NN classification accuracy with $k=1$, the 
retrieval experiments already  give an idea of its behavior (see above).

\begin{table}[t]
\small
\begin{center}
\begin{tabular}{|c|cccccc|}
\hline
Clustering (AMI) & NIST  & MARG & CLEF-IP & IH1 & IH2  & IH3 \\
\hline
CNN-A-p5 & \bf{100} & 7.9 & 31.8  & 73.6  & \bf{73.7} & 59.5 \\
\hline
CNN-G-i4a & 98.8  & 8.5  & 38.5 & 73.6  & 71.5  & \bf{65.4} \\
CNN-G-i4e & \bf{100} &  \bf{8.9}  &  \bf{40.2} & \bf{78.4}  & 66.8  & 59.3 \\
 \hline
\end{tabular}
\caption{Clustering evaluated using adjusted mutual information (AMI) for  CNN activation features that yielded best retrieval performances.}
\label{tab:CNNAMI}
\end{center}
\end{table}

\begin{table}[t]
\small
\begin{center}
\begin{tabular}{|c|cccccc|}
\hline
Classification (OA) & NIST  & MARG & CLEF-IP & IH1 & IH2  & IH3 \\
\hline
CNN-A-p5 & \bf{100} & \bf{65.7} & 75.0  & 94.3 & \bf{93.6} & 91.3 \\
\hline
CNN-G-i4a & \bf{100} & 63.3  & \bf{75.8} & 94.0 & 92.6 & 93.6 \\
CNN-G-i4e & \bf{100} & 60.4  & 74.1 & 94.5 & 92.6 & \bf{93.8} \\
\hline
\end{tabular}
\caption{NCM-based classification evaluated
  by overall accuracy (OA) for CNN activation features that yielded best retrieval performances.}
\label{tab:CNNNCM}
\end{center}
\end{table}

In what follows, we first explore the best performing models and parameter configurations of each feature type 
(shallow, deep, and hybrid), then we present overall
comparisons, and we finally discuss the results for each dataset individually.

\subsubsection{CNN features}

It is very common to use the activation of a CNN model  
trained on a dataset as ``off-the-shelf'' features for another dataset~\cite{donahue2014decaf}.
In this section, we compare activation features extracted from several activation layers, for both CNN
architectures (see details in Sections \ref{sec:cnn} and \ref{sec:training2}).
In the case of AlexNet (CNN-A), we consider the 3 most popular layers for that task: 
pool5, fc6 and fc7, that are activation features of the last pooling layer, and
of the two fully connected layers respectively. Both fc6 and fc7 have  4096 dimensions, and
the pool5 feature is 9216-dimensional.
In the case of GoogLeNet, we consider activations from the different inception layers
i3a, i3b and i4a, i4e, i5b, which are features with their dimensions  equal to 
200704, 376320, 100352, 163072 and 50176 respectively. We also consider 
the activations from the average pooling 
layer that follows the last inception layer, denoted by p5s1, which has 1024 dimensions.

We report retrieval results in Table \ref{tab:CNNMAP}. In addition we 
also report clustering and classification results in Table~\ref{tab:CNNAMI} and Table~\ref{tab:CNNNCM} for
the activation features best performing on the retrieval task.
Based on these three tables we make the following observations.

\begin{table}[t]
\small
\begin{center}
\begin{tabular}{|c|cccccc|}
\hline
Retrieval (mAP) & NIST  & MARG & CLEF-IP & IH1 & IH2  & IH3 \\
\hline
RL & \bf{100} &  34.3 &  34.8 & 63.7 & 66.5 &  57.8  \\
RL + MLP &  \bf{100} & \ul{34.9} & \ul{36.4} & \ul{66.5} & \ul{67.4} & \ul{60.7} \\
\hline
FV256 & 99.8 &  38.2 & 43.4 & 75.6 & 64.9 &  64.0 \\
FV256 + MLP & 96.5 & 35.9 & \ul{43.6} & 74.1 & 64.0 & \ul{64.8}\\
\hline
FV16 & \bf{100} & 36.3 &  44.2 & 77.4&  68.1 & 67.7  \\
FV16 + MLP &  99.7 & 32.1 & 44.0 & 76.4 & 66.3 & 65.7\\
\hline
FV256+PCA & 99.9  & \bf{38.2}  & \bf{50.9} & \bf{79.1}  & \bf{70.6} & \bf{68.3} \\
FV256+PCA+MLP & 99.8 & 37.0 & 50.8 & 77.1 & 69.6& 66.7 \\
\hline
\end{tabular}
\caption{Retrieval task: mean average precision (mAP) obtained by different shallow descriptors on different datasets.}
\label{tab:FVMAP}
\end{center}
\end{table}

First, for all three tasks, the best results are in general obtained with inception layers i4a and
i4e of the GoogLeNet network except for the MARG and IH2 dataset where the pool5 layer of AlexNet
(CNN-A-p5) outperforms in general the results obtained with the different GoogLeNet activation
features.  This can be explained by the
fact that the latter features capture higher-level semantic information, that are
well aligned with the different categories these datasets are composed of.
On the other hand, the categories from MARG and IH2 are more
correlated with the layout than with the document semantics, and the pool5 convolutional layer of the AlexNet
(CNN-A-p5) better captures the local geometry.  Surprisingly, CNN-A-p5 outperforms significantly 
CNN-A-fc6 and CNN-A-fc7 on all datasets, not only on MARG, meaning that the latter features 
does not transfer well in the context of document images. 
One explanation might be the low number of classes (16) in the 
RVL-CDIP used to train the models.

\begin{table}[t]
\small  
\begin{center}
\begin{tabular}{|c|cccccc|}
 \hline
Clustering (AMI) & NIST  & MARG & CLEF-IP & IH1 & IH2  & IH3 \\
\hline
RL & 98.9 & 4.8 & 28.5 & 18.0 & 45.7  &   19.6 \\
RL + MLP & 99.2   & \ul{6.2} &  26.7  & \ul{58.4} & \ul{52.5} &  \ul{45.5} \\
\hline
FV256 & 99.5  & 7.7 & 30.2  & 62.3 & 43.7 & 58.8 \\
FV256 + MLP & 94.2 & 6.5 & \ul{35.6} & \ul{67.7} & \ul{45.0} & \ul{63.3} \\
\hline
FV16 & \bf{100} & 7.5  & 40.3 & 71.1 & 49.7 & 63.1  \\
FV16 + MLP & 97.7 & \bf{\ul{9.1}} & 39.1 & \ul{\bf{74.6}} & \ul{54.9}  & \bf{\ul{63.5}} \\
\hline
FV256+PCA & 99.4  & 4.8  & \bf{44.9}  & 72.5 & \bf{60.3} & 47.4 \\
FV256+PCA+MLP &  98.6  & 1.6 & 42.7 & 71.2 & 59.6 & 40.9 \\
\hline
\end{tabular}
\caption{Clustering task: adjusted mutual information (AMI) for different
  shallow descriptors on different datasets.}
\label{tab:FVAMI}
\end{center}
\end{table}

\subsubsection{Shallow and hybrid features}
 
For these experiments, we consider the RunLength descriptor with spatial pyramid (RL), the two Fisher-Vector-based descriptors with respectively 16 and 256
Gaussians (FV256 and FV16), without and with and the corresponding hybrid architecture (MLP). 
In addition, we consider the PCA-projected FV256 both as shallow feature and the activation features from its hybrid architecture.

In all cases, we select the MLP model that performs best on the RVL-CDIP validation set 
(see Section in \ref{sec:part2}) and use the activation values from the
fully connected layers as feature representations, similarly to what is usually done when using CNN models as ``off-the-shelf'' features. 
By design, all these descriptors are 4096-dimensional. When using them in our three target applications, 
we observe that in most cases the activation features corresponding to the first fully
connected layer outperform the activation features of the following layers.  
Therefore we decided to only show results obtained with the first fully connected layer.

We show results both with 
the shallow features and the corresponding hybrid features in
Table~\ref{tab:FVMAP} for retrieval, Table~\ref{tab:FVAMI} for clustering, and
Table~\ref{tab:FVNCM} for categorization. Best results per dataset are shown in bold. 

We observe that unlike for the classification task, when used in
transfer, the advantage of hybrid architectures is less obvious. To make this
easier to observe from the tables, we underline the cases where the hybrid activation
feature outperforms its corresponding shallow feature. The results are
somewhat mixed depending on features, tasks and datasets.  The activation feature of
the hybrid model learned on RL is almost always better than the original RL
feature. In the case of the FV16 and FV256 the hybrid model sometimes brings a
gain (especially on clustering results) but in other cases using directly the
shallow features performs better. If we consider the PCA reduced FV256, using the
hybrid model most often degrades the performance.

Overall, best retrieval results and most often best NCM classification accuracies
are obtained with FV256+PCA. Concerning clustering, FV16+MLP significantly 
outperforms FV256+PCA for three datasets out of six.

\begin{table}[t]
\small  
\begin{center}
\begin{tabular}{|c|cccccc|}
 \hline
Classification (OA) & NIST  & MARG & CLEF-IP & IH1 & IH2  & IH3 \\
\hline
RL & \bf{100} & 54.7 & 57.8  &  90.2 & 79.8& 83.2\\
RL + MLP & \bf{100} & 53.4 & \ul{63.1} & \ul{91.4} & 79.1 & \ul{84.9} \\
\hline
FV256 & \bf{100}  &  62.0  & 76.7 & 92.6 & 83.5 & 90.0 \\
FV256+MLP & \bf{100}  &   58.4 & 69.4 & 92.2 & 81.9 & 89.5 \\
\hline
FV16 & \bf{100} & 61.7  & 72.8 & 92.9 & 85.0 & 90.1 \\
FV16+MLP & \bf{100} &  53.7 & 64.5 & \ul{93.1} & 84.1 & 88.4 \\
\hline
FV256+PCA & \bf{100}  & \bf{64.4} & \bf{81.3} & 94.1 & \bf{86.6} & \bf{92.5}\\
FV256+PCA+MLP &\bf{100}  & 63.1 & 79.6 & \ul{\bf{94.5}} & 86.4&  \bf{92.5} \\
\hline
\end{tabular}
\caption{NCM-based classification: overall accuracy (OA) obtained by different shallow
  and hybrid descriptors on different datasets.}
\label{tab:FVNCM}
\end{center}
\end{table}


\subsubsection{Comparing shallow and deep features}

Finally, in Table \ref{tab:ALL}, we summarize all the best results obtained with shallow, deep, and hybrid features and 
analyze them dataset per dataset.

\begin{table}[t]
\small
\begin{center}
\begin{tabular}{|c|cccccc|}
\hline
Retrieval (mAP) & NIST  & MARG & CLEF-IP & IH1 & IH2  & IH3 \\
\hline
CNN-A-p5 & \bf{100} & \bf{38.7} & 40.9 & 75.1 & 78.8  & 67.7 \\
CNN-G-i4a & \bf{100} & 36.4 & 42.4 & 78.5 & \bf{81.8} & 73.4 \\
CNN-G-i4e & \bf{100} & 36.2 & 43.9  & 78.0& 80.3 & \bf{73.8} \\
\hline
FV16 + MLP &  99.7 & 32.1 & 44.0 & 76.4 & 66.3 & 65.7\\
FV256+PCA & 99.9  & 38.2  & \bf{50.9} & \bf{79.1}  & 70.6 & 68.3 \\
\hline
\hline
Clustering (AMI) & NIST  & MARG & CLEF-IP & IH1 & IH2  & IH3 \\
\hline
CNN-A-p5 & \bf{100} & 7.9 & 31.8  & 73.6  & \bf{73.7} & 59.5 \\
CNN-G-i4a & 98.8  & 8.5  & 38.5 & 73.6  & 71.5  & \bf{65.4} \\
CNN-G-i4e & \bf{100} &  8.9  &  40.2 & \bf{78.4}  & 66.8  & 59.3 \\
\hline
FV16 + MLP & 97.7 & \bf{9.1} & 39.1  & 74.6 & 54.9  & 63.5 \\
FV256+PCA & 99.4  & 4.8  & \bf{44.9}  & 72.5 & 60.3 & 47.4 \\
\hline
\hline
Classification (OA) & NIST  & MARG & CLEF-IP & IH1 & IH2  & IH3 \\
\hline
CNN-A-p5 & \bf{100} & \bf{65.7} & 75.0  & 94.3 & \bf{93.6} & 91.3 \\
CNN-G-i4a & \bf{100} & 63.3  & 75.8 & 94.0 & 92.6 & 93.6 \\
CNN-G-i4e & \bf{100} & 60.4  & 74.1 & \bf{94.5} & 92.6 & \bf{93.8} \\
\hline
FV16 + MLP & \bf{100} &  53.7 & 64.5 & 93.1 & 84.1 & 88.4 \\
FV256+PCA &\bf{100}  & 64.4 & \bf{81.3} & 94.1 & 86.6 & 92.5\\
\hline
\end{tabular}
\caption{Summary table that compares best performing variants of the different
  descriptors for retrieval, clustering and classification, on different datasets.}
\label{tab:ALL}
\end{center}
\end{table}

\paragraph{\bf NIST} This dataset is much easier than the other ones, as the appearance is
very consistent within a given category ,and categories are well-aligned with specific templates. 
Consequently, most methods perform really well on all tasks.

\begin{figure}[t]
\centering
\includegraphics[width=1\linewidth]{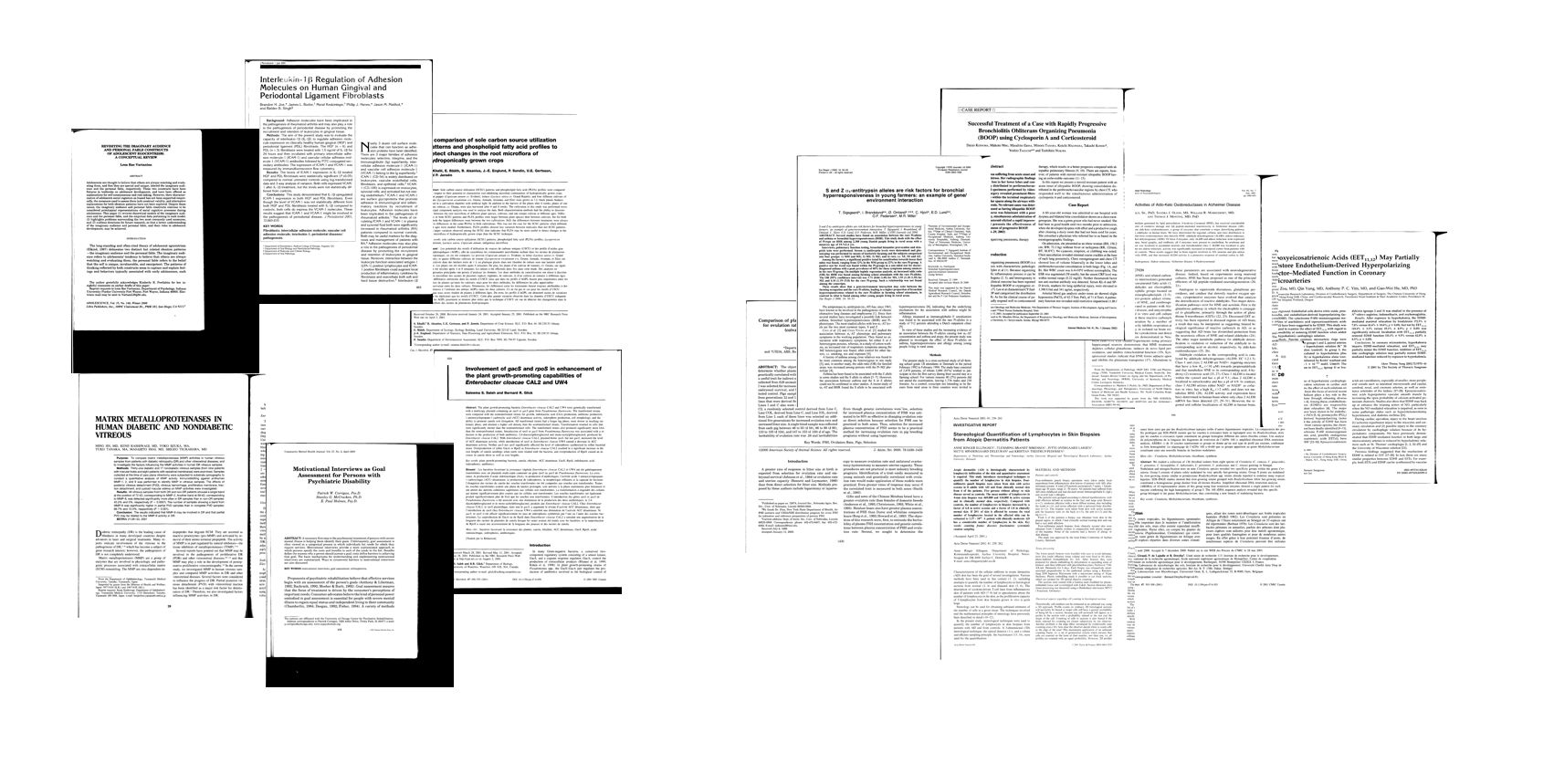}
\caption{Illustration of the MARG dataset. Left: visually different documents
  that all share the same category label. Right: documents from different
  categories grouped automatically in the same cluster.}
\label{fig:MARGClustEx}
\end{figure}

\paragraph{\bf MARG} This dataset is much more challenging as category labels were
defined based on specific aspects of the document layout (such as the presence and the location
of the title, affiliation or the abstract), while other aspects of the layout are totally ignored (\eg the
number of columns in the document). Consequently, there is large intra-class
variation and visually similar documents can belong to different categories.
This is illustrated in Figure \ref{fig:MARGClustEx}. The left part of the
figure displays visually dissimilar documents from the same category, while the
right part displays a cluster of visually similar documents that belong to
different categories (each document represents one of the classes).
This could explain the very low clustering results obtained,  independently of
the visual feature used. Regarding the retrieval and NCM classification tasks, best
 results are obtained with pool5 activation features from AlexNet (CNN-A-p5), however the performance 
 obtained with FV256+PCA are close to these results and better than the results obtained with GoogLeNet 
 activation features.
 
\begin{figure}
\centering
\includegraphics[width=0.9\linewidth]{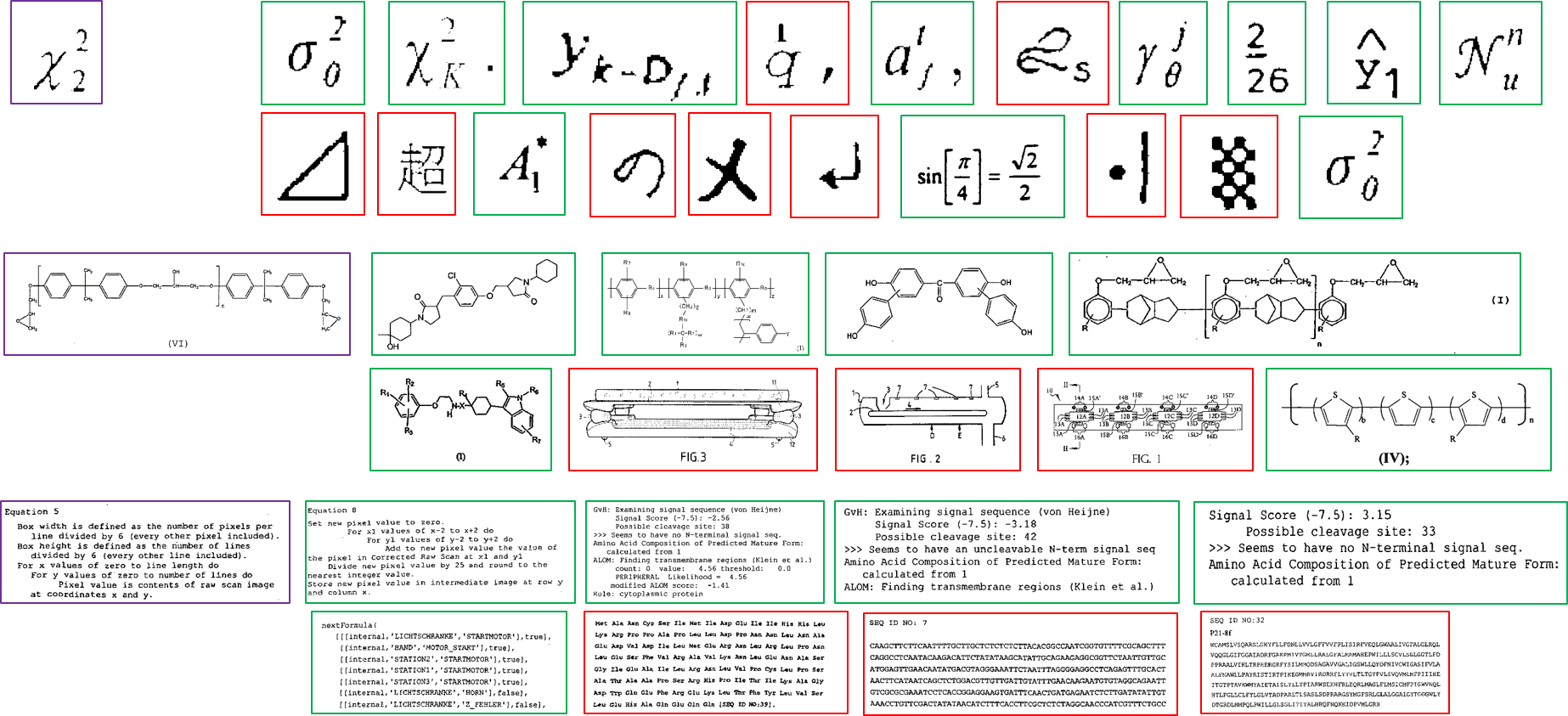}
\caption{Randomly selected query examples from Clef-IP dataset (with
violet borders at left). For each query, top results with FV256+PCA on
the upper row and below top results obtained with CNN-G-i4e. Relevant
results are shown with green borders and irrelevant ones with red borders
(best viewed in color).}
\label{fig:clefret1}
\end{figure}

\begin{figure}
\centering
\includegraphics[width=0.95\linewidth]{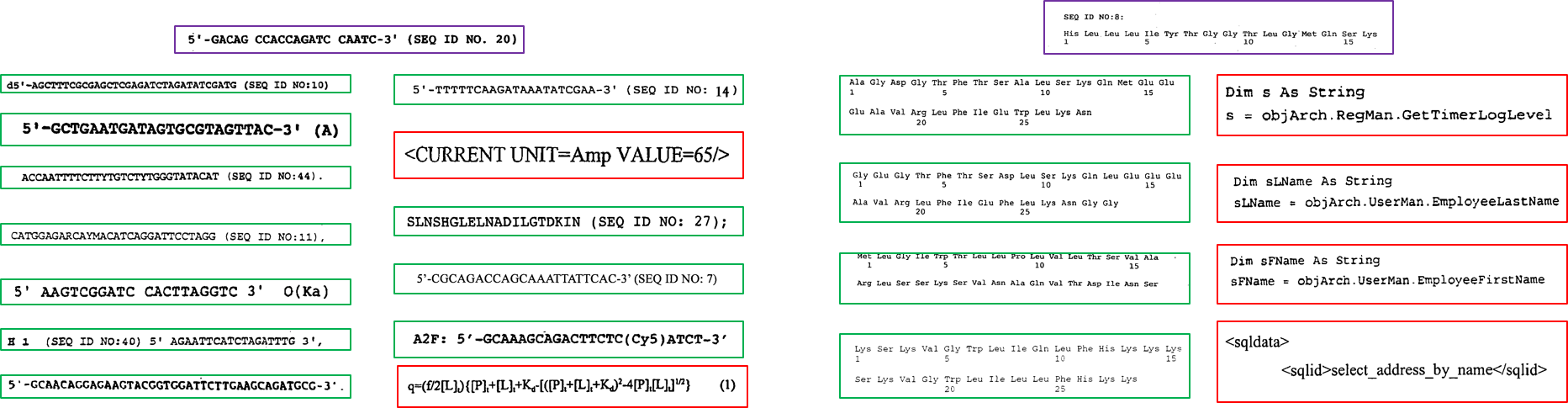}
\caption{Randomly selected query examples from Clef-IP dataset (in violet on top). 
For each query, top results with FV256+PCA on left 
and with  CNN-G-i4e on right. Relevant results are shown with green border 
and irrelevant ones with red borders (best viewed in color).}
\label{fig:clefret2}
\end{figure}

\paragraph{\bf Clef-IP} This dataset departs from the others in two aspects. First, the size and aspect 
ratio of the images varies a lot, which might have a strong impact on CNN
representations that use a fixed size and aspect ratio as input. Second, there
is a very large intra-class variability and the document layout has small or
even no importance in the category definition (see examples in Figure 
\ref{fig:benchmarkdataex}). This might explain why FV256+PCA outperforms by a
large margin CNN activation features; the former are explicitly designed to work
with geometry-less bags of local features, and consequently they better capture local
information disregarding its position (see \eg flowchart components or
mathematical symbols in formulas from Figure \ref{fig:benchmarkdataex}).
Qualitative results can be seen in Figures~\ref{fig:clefret1} and  \ref{fig:clefret2}. These figures display
randomly chosen queries and the corresponding top retrieval results obtained with FV256+PCA and 
CNN-G-i4e respectively.

\paragraph{\bf IH1} This dataset is probably the one most similar to the RVL-CDIP
dataset, on which the feature representations have been trained. Indeed, both
datasets share classes, such as \textit{invoices}, \textit{contracts}, etc. However, the IH1
dataset also consists of sub-classes (\eg \textit{invoice type 1} and
\textit{invoice type 2}). On this dataset, for the classification task,  the different
methods obtain similar accuracies, CNN-G-i4e features being the best. 
This feature yields also the best clustering performance
but is outperformed on the retrieval task by FV256+PCA. The relatively 
good and similar performances obtained with the CNN and hybrid activation features is probably due to 
the closeness between the classes and images of the  RVL-CDIP dataset, used to train 
the models, and the IH1 dataset.

\paragraph{\bf IH2} This is a small fine-grained dataset (72 categories) where the
document layout (\eg "\textit{page with two tables, one on the top and one on the bottom}"),
plays a crucial role in the category definition. This property
seems to have been better captured by CNN activation features that keep geometric information
compared to FV256+PCA or FV16+MLP that are less dependent on the layout. The importance 
to capture the geometry for this dataset can be seen also by a deeper analyses of the
Tables~\ref{tab:FVMAP},~\ref{tab:FVAMI} and~\ref{tab:FVNCM} where 
we can see that FV16 (with its spatial grid) outperforms FV256. It can also be seen on retrieval and clustering
where even RL (with 5 layered spatial pyramid) outperforms FV256. In summary, on this dataset, 
there is no obvious best performing feature, CNN-A-p5 performs the best for clustering and NCM classification, but it
is outperformed by both CNN-G-i4a  and CNN-G-i4e on the retrieval task.

\begin{figure}
\centering
\includegraphics[width=0.85\linewidth]{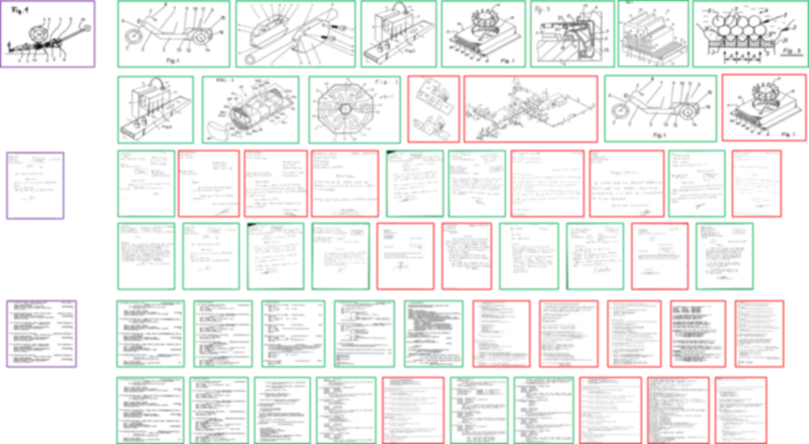}
\caption{Randomly selected examples from IH3 (at left). Top results with FV256+PCA (upper row)
and with CNN-G-i4e (below). (The images  were intentionally
blurred to keep the actual content of the documents confidential.}
\label{fig:IH3ret}
\end{figure}

\paragraph{\bf IH3} The last in-house dataset contains 63 fine-grained categories such as diverse \textit{forms, 
invoices, contracts} where each form/contract/invoice coming from a different customer corresponds to a different class.
It can be seen as a mix between NIST (as some classes are variations of templates) and IH1 (other classes are more generic 
 with intra-class variations, and we have also sub-classes for several of them). On this dataset best or close 
 to best results were obtained with the  CNN-G-i4e activation features of the  GoogLeNet.  
 We show some retrieval examples in Figure~\ref{fig:IH3ret} where we 
compare the top results for this feature with the top results obtained with FV256+PCA.
Note that the class label differences often come from the fact that the document belongs to different customers, which explains that 
while most retrieved documents are of the same generic type as the query (e.g. drawing, handwritten letter, printed code) 
not all of them are considered as relevant to the query (provided by different customers they belong to different classes).

\section{Conclusions}
\label{sec:ccl}

This paper proposes a detailed benchmark that compares three types of document image representation: so-called shallow 
features, such as the RunLength and the
Fisher-Vector descriptors, deep features based on Convolutional Neural Networks, 
and features extracted from hybrid architectures that take inspiration from the
two previous ones. Our benchmark first compares these features on a 
classification task where the training and testing sets belong to the same domain. It also
compares these features when used to represent documents from other domains, 
for three different tasks, in order to quantify how much these different document
image representations generalize across datasets and tasks.

We observed that without domain shift, Convolutional Neural Network features 
perform better than shallow and hybrid features, closely followed by hybrid architectures that perform
almost as well for a fraction of the training cost. 
This had already been observed for natural images, and we confirmed this observation for document images.

In presence of a domain shift, the story changes quite significantly. 
Independently of the targeted task (we considered retrieval, clustering, and
classification), the hybrid architectures do not transfer well in general across datasets.  
Instead, deep or shallow features are the best, depending on the dataset
specificities. On one hand, Convolutional Neural Networks seems to perform the best for target datasets 
that are not too different from the source dataset, and
for datasets for which the global layout is important. 
On the other hand, PCA reduced FVs appears to better deal with strong aspect-ratio changes and very large intra-class
variability on the document layout.

\bibliographystyle{myplain}
\bibliography{docimgs}

\end{document}